\documentclass[letterpaper, 10pt, conference]{ieeeconf/ieeeconf}    

\makeatletter
\let\NAT@parse\undefined
\makeatother

\usepackage[numbers,sectionbib,sort&compress]{natbib}

\let\labelindent\relax
\usepackage{enumitem}

\usepackage{bm}
\usepackage{gensymb}
\usepackage{graphicx}
\usepackage{amsmath}
\usepackage{algorithm}
\usepackage[font=small,skip=0pt]{subcaption}
\usepackage{amsfonts}
\usepackage{siunitx}
\usepackage{booktabs}
\usepackage[font=small]{caption}
\usepackage[export]{adjustbox}
\usepackage{sidecap} \sidecaptionvpos{figure}{c}
\usepackage{multirow}
\usepackage{algorithmic}
\usepackage{comment}

\DeclareMathOperator*{\argmax}{\arg\!\max}

\usepackage{hyperref}
\hypersetup{colorlinks,breaklinks,
linkcolor=[rgb]{0.5,0.,0.},
citecolor=[rgb]{0.000,0.427,0.173},
urlcolor=[rgb]{0.031,0.318,0.612}}

\usepackage[printonlyused,withpage,nolist,nohyperlinks]{acronym}
\begin{acronym}

\acro{2D}{two-dimensional}

\acro{3D}{three-dimensional}

\acro{AHRS}{attitude and heading reference system}

\acro{AUV}{autonomous underwater vehicle}

\acro{CPP}{Chinese Postman Problem}

\acro{DoF}{degree of freedom}
\acrodefplural{DoF}[DoFs]{degrees of freedom}

\acro{DVL}{Doppler velocity log}

\acro{FSM}{finite state machine}

\acro{IMU}{inertial measurement unit}

\acro{LBL}{Long Baseline}

\acro{MCM}{mine countermeasures}

\acro{MDP}{Markov decision process}
\acrodefplural{MDP}[MDPs]{Markov decision processes}

\acro{POMDP}{partially observable Markov decision process}
\acrodefplural{POMDP}[POMDPs]{partially observable Markov decision processes}

\acro{PRM}{Probabilistic Roadmap}
\acrodefplural{PRM}[PRM]{Probabilistic Roadmaps}

\acro{ROI}{region of interest}
\acrodefplural{ROI}[ROIs]{regions of interest}

\acro{ROS}{Robot Operating System}

\acro{ROV}{remotely operated vehicle}

\acro{RRT}{Rapidly-exploring Random Tree}
\acrodefplural{RRT}[RRTs]{Rapidly-exploring Random Trees}

\acro{SLAM}{simultaneous localization and mapping}

\acro{SSE}{sum of squared errors}

\acro{STOMP}{Stochastic Trajectory Optimization for Motion Planning}

\acro{TRN}{Terrain-Relative Navigation}

\acro{UAV}{unmanned aerial vehicle}

\acro{USBL}{Ultra-Short Baseline}

\acro{IPP}{informative path planning}

\acro{FoV}{field of view}
\acrodefplural{FoV}[FoVs]{fields of view}

\acro{CDF}{cumulative distribution function}

\acro{ML}{maximum likelihood}

\acro{RMSE}{Root Mean Squared Error}
\acro{MLL}{Mean Log Loss}

\acro{GP}{Gaussian Process}
\acrodefplural{GP}[GPs]{Gaussian processes}

\acro{KF}{Kalman Filter}

\acro{IP}{Interior Point}
\acro{BO}{Bayesian Optimization}

\acro{GPS}{Global Positioning System}

\acro{PDM}{probability density map}

\end{acronym}

\IEEEoverridecommandlockouts                           
\title{\LARGE \bf
Multi-agent Time-based Decision-making \\ for the Search and Action Problem
}

\author{
    Takahiro Miki, Marija Popovi\'{c}, Abel Gawel, Gregory Hitz and Roland Siegwart
\thanks{Authors are with the Autonomous Systems Lab., ETH Zurich, Switzerland.  \tt\small takahiro.miki1992@gmail.com,\{mpopovic, gawela, hitzg, rsiegwart\}@ethz.ch}.}

\begin{document}

\maketitle
\thispagestyle{empty}
\pagestyle{empty}

\begin{abstract}
Many robotic applications, such as search-and-rescue,
require multiple agents to search for and perform actions on targets.
However, such missions present several 
challenges,
including 
cooperative exploration, task selection and allocation,
time limitations, and computational complexity.
To address this,
we propose a decentralized multi-agent decision-making framework
for the search and action problem with time constraints.
The main idea is to treat time as an allocated budget
in a setting where each agent action incurs a time cost and yields a certain reward.
Our approach leverages probabilistic reasoning
to make near-optimal 
decisions leading to maximized reward.
We evaluate our method in the search, pick, and place scenario
    of the Mohamed Bin Zayed International Robotics Challenge (MBZIRC),
by using a probability density map and reward prediction function
to assess actions.
Extensive simulations show that our algorithm outperforms benchmark strategies,
and we demonstrate system integration
in a Gazebo-based environment, validating the framework's readiness for field application.

\end{abstract}

\section{INTRODUCTION}
\label{sec:introduction}

Thanks to ongoing developments in sensing and processing technologies,
mobile robots are becoming more capable of working in dynamic and challenging environments
in many practical applications, such as search-and-rescue~\citep{rudol2008human,murphy2008search},
multi-robot exploration~\citep{kingston2016automated,jin2003cooperative,Maza2007multiple} and
terrain monitoring~\citep{Popovic2017IROS}.
In many multi-agent applications, however, efficient task allocation remains a field of open research.
To perform fully autonomous missions,
algorithms for agent cooperation and efficient area search are required.

To address this,
our work focuses on cooperative planning strategies within the context of the
Mohamed Bin Zayed International Robotics Challenge (MBZIRC)\footnote{\url{http://www.mbzirc.com/}}~\citep{bahnemann2017mbzirc}.
In one stage of this competition (Challenge 3),
a team of \acp{UAV} is required to collaboratively
search, pick, and place a set of static and moving objects,
gaining points for each one collected.
This task poses several challenges:

\begin{itemize}
\item coordinating multiple \acp{UAV} to explore the field,
\item tracking moving targets efficiently,
\item trading off exploration to find new objects with picking them up to score points,
\item making decisions based on the time limitation of a task.
\end{itemize}

A key aspect of such missions is that the timing to execute actions should be considered, given the targets found.
In the MBZIRC, for instance, a \ac{UAV} can greedily pick up an object to attain a certain score.
However, 
it could be better to  invest in exploration to find a more valuable object nearby.
The optimal decision here includes several aspects, and differs with the time remaining until mission completion.
With stricter time limits, exploration becomes riskier and acting greedily might be preferred.
The exploration-exploitation trade-off must thus be addressed
while accounting for the fact that optimal decisions differ at various stages of the mission.
\begin{figure}[tbp]
  \begin{center}
    \includegraphics[width=\columnwidth]{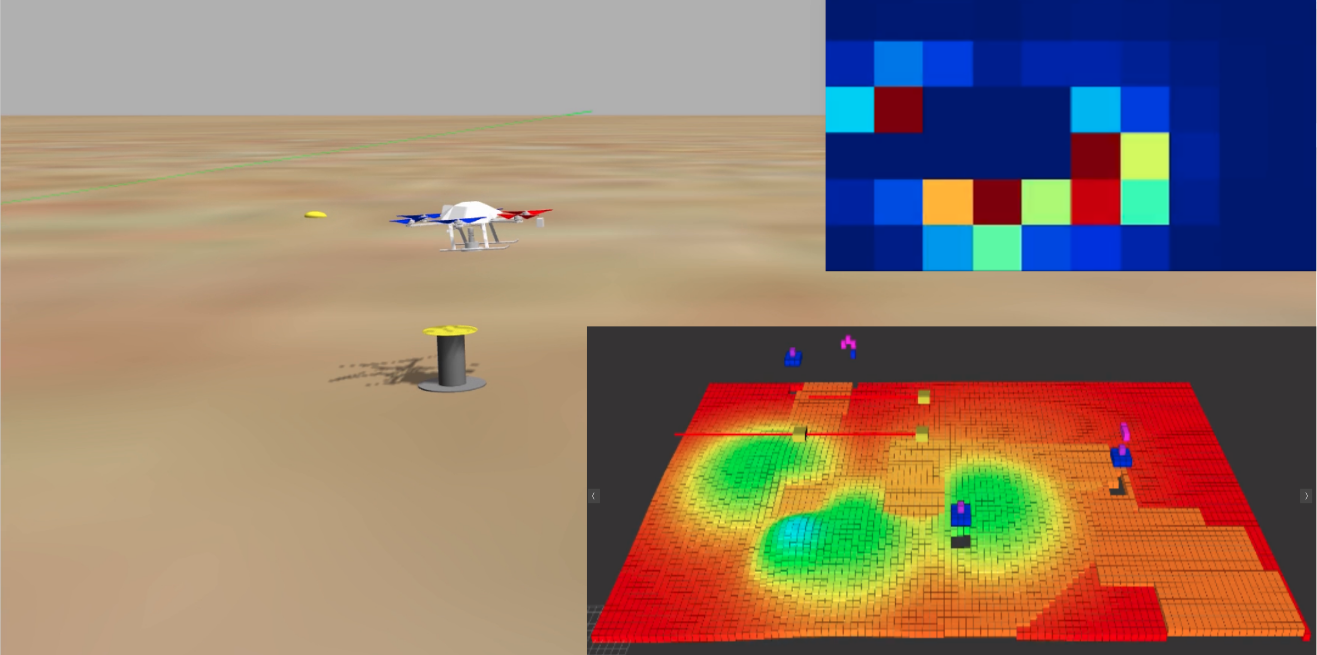}
    \caption{Our decision-making algorithm applied to the search, pick, and place scenario of the MBZIRC
    in a 3D simulation.
    We use a \ac{PDM} (bottom-right) to plan exploratory paths on a grid representing the sum of expected scores to be found,
    calculated from the \ac{PDM} (upper-right).
    By predicting the value of search and action decisions using the \ac{PDM},
    three \acp{UAV} coordinate implicitly to maximize the value of successful deliveries given the time constraints.
    }
    \label{fig:field}
  \end{center}
\end{figure}

To tackle these issues,
this paper introduces a multi-agent decision-making algorithm which
accounts for (i) target movement,
(ii) mission time constraints,
and (iii) variable target rewards.
The main idea is to treat time as a budget used by the agents.
At a given time, all possible actions consume some budget
while yielding a certain reward.
Each agent has an initial budget specified as the mission time limit,
and must choose the sequence of actions maximizing the final reward.
To evaluate actions, future rewards are predicted by planning on a probabilistic map.
A key aspect of our approach is that all agents can operate not fully synchronously.
Moreover, by using implicit coordination~\citep{hollinger2008proofs},
it does not suffer from computational growth with the number of agents,
making it applicable to real-time scenarios.

The contributions of this paper are:
\begin{enumerate}
\raggedright
\item A multi-agent decision-making framework which:
\begin{itemize}
\item is decentralized and real-time with near-optimality guarantees,
\item considers a fixed time budget,
\item addresses the search and action problem without using a trade-off parameter.
\end{itemize}
\item The validation of our framework on Challenge 3 of the MBZIRC,
with an evaluation against alternative strategies.
\end{enumerate}

While our framework is motivated by the MBZIRC,
it can be used in any multi-agent search and action
scenario.

The remainder of this paper is organized as follows.
Section 2 describes related work.
We formulate our proposed method as a general search and action problem in Section 3.
Sections 4 and 5 detail our experimental set-up and results.
In Section 6, we conclude with a view towards future work.

\section{RELATED WORK}
Significant work has been done in the field of decision-making for the search and action problem.
This section provides a brief overview.
We first discuss searching for targets only before also considering executing tasks on them.

\subsection{Coverage Methods}
Pure exploration tasks are typically addressed using complete coverage methods.
The aim is to generate an obstacle-free path passing through all points in an area,
usually by decomposing the problem space~\citep{Galceran2013survey}.
Based on these ideas,
\citet{Maza2007multiple} proposed a sweep-based approach for target search using multiple \acp{UAV}. 
However, their strategies are not applicable in our setup due to the presence of moving targets,
which can enter areas already explored.

\subsection{Search and Pursuit-evasion}
There are many search algorithms which allow for target movement.
\citet{isler2011search} categorized autonomous search problems using several features, 
dividing them into two categories: (i) adversarial pursuit-evasion games, and (ii) probabilistic search.

In a pursuit-evasion game, pursuers try to capture evaders as evaders try to avoid capture.
One graph-based example of this game is {\it cops-and-robbers}~\citep{nowakowski1983vertex,aigner1984game},
which assumes that players have perfect knowledge of each other's positions.
Similarly, \citet{adler2003randomized} studied {\it hunter-and-rabbit},
where players do not know each other's positions unless found on the same vertex.
Their search method aims to maximize worst-case performance
so that the targets can be found regardless of how they move.
These deterministic methods, however,
do not cater for practical sensing limitations.

In contrast, probabilistic search methods try to maximize probability of detection or minimize expected detection time,
assuming that the targets' and searchers' movements are independent of each other.
Like \citet{wong2005multi}, we follow a Bayesian approach to compute the probability density of targets
and use this to coordinate multiple searchers.

Formally, general search problems can be formulated as a 
Partially Observable Markov Decision Process (POMDP) with hidden states.
However, real-time planning within such frameworks is difficult 
because of the exponential growth in computation with the number of searchers,
even with near-optimal 
solvers~\citep{smith2007probabilistic,kurniawati2008sarsop}.
To address this, \citet{hollinger2008proofs} propose implicit coordination with receding-horizon planning
to achieve near-optimality
while scaling only linearly with the number of teams.

The above methods assume that
all agents act simultaneously, 
i.e., using a synchronization system.
Alleviating this requirement, \citet{messias2013asynchronous}
study a multi-agent POMDP for asynchronous execution.


\subsection{Combined Search and Action}
While the strategies above are suitable for efficient target search,
they do not take into account any subsequent tasks on them (e.g., object pickup and delivery).
\citet{hollinger2009combining} were first to examine both aspects in the context of robotics.
In their work, the aim is to minimize the combined search and action time with targets found in an environment using finite-horizon plans.
However, a key assumption is that actions are executed as soon as targets are found.
We leverage similar principles,
but also allow agents to choose between search and action
given the time budget constraints.

\section{PROPOSED APPROACH}
In this section, we present our planning approach for the general search and action problem.
By maximizing the attainable reward,
our strategy aims to efficiently decide whether to explore an environment or to execute actions
given the targets found so far.
We begin by defining the problem, then outline the details of our algorithm.

\subsection{Problem Setup}
\label{sec:problem_setup}
Our problem involves finding tasks associated with static and moving targets
and performing actions to complete them within a specified time limit.

We consider a graph-like grid environment with edge-connected cells.
Each edge is weighted by a distance
and an agent moves to detect tasks in its current cell.
The agent searches for tasks associated with multiple targets,
which can be either static or moving.
The motion of moving targets is non-adversarial,
i.e., it does not change with the searchers' behaviour.
Each target task can be executed through an action,
which incurs a time cost and yields a certain reward.

The objective is to obtain the highest possible reward within the time limit with multiple search agents.
To achieve this, they must explore the field, locate the tasks, and execute the associated actions cooperatively. 

\subsection{Algorithm Design}
The main aspects of the algorithm are: (i) decision-making for search vs. task execution,
(ii) evaluation of action candidates,
(iii) consideration of time constraints, and
(iv) cooperation of multiple agents.

To solve this problem, we extend the Multi-robot Efficient Search Path Planning (MESPP) algorithm proposed by~\citet{hollinger2008proofs}.
The agents try to maximize a reward function within a finite horizon.
Then, they cooperate implicitly by executing the actions
in a sequential manner.
Efficiency is provably within $(1 - 1 / e) \approx 0.632$ of the optimum when the optimization function is non-decreasing and submodular~\citep{krause2012submodular}.
To apply this for our problem, we define an optimization function with these properties
that accounts for the remaining time at a particular stage in the mission. 
Our function is derived in three steps:
\begin{enumerate}
\item Definition of a function which predicts the total reward within the time limit.
\item Formulation of an action-evaluation function, and demonstration of a decision-making algorithm which employs this function.
\item Derivation of the non-decreasing submodular optimization function for multiple agents.
\end{enumerate}



\subsubsection{Reward prediction function}
\label{sec:RewardFunction}
This function forecasts the total reward attainable within the time limit. 
We assume that $m$ agents have a set of $n$ already found tasks $\mathcal{T}$.
Each task $i$ is associated with an execution time (cost) $c_i$ and a reward $r_i$. 
With this information, the total reward can be predicted by choosing tasks such that the total time cost does not exceed a given constraint.
This reward prediction function is represented as $\mathrm{J}(\mathcal{T}, t)$,
where $j$ indexes the time budget remaining in $t$ for a particular agent:
\begin{eqnarray}
\mathrm{J}(\mathcal{T}, t) = \max \sum_{j=1}^m \sum_{i=1}^n r_i x_{ij}\text{,} \label{E:reward_prediction}
\end{eqnarray}
with 
\begin{eqnarray}
\begin{array}{ll}
  \sum_{i=1}^n c_i x_{ij} \leq t_j, & j \in  \{1, \dots, m\}\\
  \sum_{j=1}^m x_{ij} \leq 1, & i \in  \{1, \dots, n\}
\end{array}\\
	x_{ij} = \left\{
	\begin{array}{ll}
		1 & \text{if task $i$ is assigned to agent $j$},\\
		0 & \text{otherwise}.\\
	\end{array}
	\right.
\end{eqnarray}





\subsubsection{Action-evaluation function and decision-making algorithm}
The action-evaluation function assesses exploratory actions as the probability of the predicted reward increases.
An exploratory action $a$ consuming time $c_a$ could increase or decrease the total reward,
depending on whether or not it leads to finding new, more valuable, tasks.

The utility of exploration depends on the set of already found tasks $\mathcal{T}$, remaining time budget $t$, and probability of finding new tasks. 
By accounting for these effects, the action-evaluation function $\mathrm{R}_\mathcal{T}(a)$ computes the reward increase:
\begin{eqnarray}
\mathrm{R}_\mathcal{T}(a) = p(\mathcal{T}_a | a) \left[\mathrm{J}(\mathcal{T} + \mathcal{T}_a, t - c_a) - \mathrm{J}(\mathcal{T}, t) \right] \text{,} \label{E:reward_function}
\end{eqnarray}
where
$\mathcal{T}_a$ is the set of tasks findable by performing action $a$ (including the case that no task is found),
$p(\mathcal{T}_a | a)$ represents the probability of finding the set of tasks $\mathcal{T}_a$ with action $a$, and
$\mathrm{J}(\mathcal{T} + \mathcal{T}_a, t - c_a)$ denotes the total reward with new tasks
after deducting the cost $c_a$ from the agent's budget $t_j$.

Algorithm~\ref{alg:task_exe} shows our logic for determining the best action $a^*$ based on the expected reward increase.
Eq.~\ref{E:reward_function} is used to evaluate all exploratory actions (Lines~1-3) and pick the best one (Line~4).
If the reward function decreases with $a^*$,
a known task is executed instead (Lines~5-7).
\begin{algorithm}[ht]
\caption{\textsc{select\_action} procedure}         
\label{alg:task_exe}                          
\begin{algorithmic}[1]
\REQUIRE Found tasks $\mathcal{T}$, time budget $t$
\ENSURE Highest-value action $a^*$
\FOR{All exploratory actions $a$}
\STATE Calculate $\mathrm{R}_\mathcal{T}(a)$ using Eq. \ref{E:reward_function}.
\ENDFOR
\STATE $a^* \leftarrow  \underset{a}{\argmax}\mathrm{R}_\mathcal{T}(a)$
\IF{$\mathrm{R}_\mathcal{T}(a^*) < 0$}
\STATE \textsc{execute\_task()}
\ENDIF
\end{algorithmic}
\end{algorithm}

\subsubsection{Optimization function}
Finally, we define an optimization function $\mathrm{F}(A)$ for executing a sequence of actions $A$:
\begin{eqnarray}
\mathrm{F}(A) = \mathrm{J}(\mathcal{T}_A, t_0)\text{,} \label{E:optimization_function}
\end{eqnarray}
where $\mathcal{T}_A$ is the set of tasks found along actions $A$ and $t_0$ is the initial time limit.

The key idea is that executing Algorithm~\ref{alg:task_exe} is the same as maximizing $\mathrm{F}(A\cup a) - \mathrm{F}(A)$ for a single action $a$.
Therefore, if Eq.~\ref{E:optimization_function} is nondecreasing and submodular\footnote{A proof of the properties required for near-optimality is presented in the Appendix.},
a sequential decision-making procedure at each replanning stage gives near-optimal solutions with multiple agents (Algorithm~\ref{alg:sequential}).
After one agent plans, the chosen action is taken into account through the reward function, allowing the others to plan based on its decision. 
A key benefit of this approach is that its complexity increases only linearly with the number of agents,
making it feasible for real-time planning.

\begin{algorithm}[htbp]                      
\caption{\textsc{replan\_actions} procedure}         
\label{alg:sequential}                          
\begin{algorithmic}
\REQUIRE Found tasks $\mathcal{T}$, time budget $t$
\WHILE{$t > 0$}
\STATE \textsc{select\_action}($\mathcal{T}$, $t$)
\STATE Update $p(i|a)$, $t$, and $\mathcal{T}$.
\ENDWHILE
\end{algorithmic}
\end{algorithm}

\section{MBZIRC APPLICATION}
This section overviews the MBZIRC task (Challenge 3) motivating our general decision-making approach.
Then, we describe how its elements are adapted for this scenario.

\label{sec:mbzirc}
\subsection{Challenge Description}
Challenge 3 of the MBZIRC takes place in an $100$\,m~$\times$~$60$\,m field containing three types of objects.
Each object has a different score, distinguishable by color.
Using three \acp{UAV}, the objective is to explore the field for objects,
then pick up and drop them in a designated box to score points.
Points are only obtained upon successful delivery to the box.
The three object classes are:
\begin{itemize}
 \item $Static$. Does not move, has several point categories.
 \item $Moving$. Moves randomly, has a higher point category.
 \item $Large$. Does not move, requires \ac{UAV} collaboration to pick up\footnote{For simplicity, this work does not consider large objects.
We note that they can be easily included in the current framework using simple cooperative logic.}.
\end{itemize}

The field contains ten static, ten moving, and three large objects.
The dropping box is located in the middle of the field
and the three \acp{UAV} start from the same point.
The time limit for the mission is $20$\,mins.

\subsection{Action-evaluation Function}
To show the flexibility of our framework,
this section describes how
it can be adapted to the MBZIRC setup.
We specify the task set
and define its probability density for exploration.
Based on these ideas,
we formulate the reward prediction function (Eq.~\ref{E:reward_function}) for decision-making.

\subsubsection{Task definitions} \label{ss:task_definitions}
We associate a new task to each object in the arena.
The cost of a task $c_i$ is the time taken to complete the pick-up and drop-off action,
calculated as:
\begin{eqnarray}
\label{func: pickup_cost}
c_i &=& t_{approach} + t_{pick} + t_{transfer} + t_{drop}\text{,}
\end{eqnarray}
where $t_{approach}$ is the time for a \ac{UAV} to move to the object from its current position,
$t_{pick}$ is the time to pick up the object given its type,
and $t_{transfer}$ and $t_{drop}$ are the times to transfer the object to and deposit it in the dropping box.

The reward of a task $r_i$ is simply the score points that are obtained upon the successful delivery of an object.

\subsubsection{Probability of finding new tasks}
The probability of finding new tasks (objects)
is expressed using a Probability Density Map (\ac{PDM}),
updated through a Bayesian filtering procedure.
The \ac{PDM} is created by discretizing the arena area into a grid,
where each cell represents a state $x$.
At time-step $t$,
the filter calculates the belief distribution $bel$ from measurement and control data
using the equations~\citep{thrun2002probabilistic}:
\begin{eqnarray}
\overline{bel}(x_i) &=& \int p(x_t | u_t, x_{t-1})bel(x_{t-1})dx\text{,} \label{E:bayes1} \\
bel(x_t) &=& \eta p(z_t | x_t)\overline{bel}(x_t)\text{,} \label{E:bayes2}
\end{eqnarray}
where $p(x_t | u_t, x_{t-1})$ is the transition probability between successive states given the control input $u_t$, 
$p(z_t|x_t)$ is the measurement model for observation $z_t$,
and $\eta$ is a normalization constant.


To handle multiple tasks,
seperate \acp{PDM} are stored for each object
and updated using Eqs.~\ref{E:bayes1} and~\ref{E:bayes2}.
Note that $u_t$ is neglected
since the \acp{UAV} have no control over the objects' motion.
Static objects
maintain a constant probability:
\begin{equation}
p(x_t | u_t, x_{t-1}) = \begin{cases}
1 & \text{if } x_t = x_{t-1}\text{,} \\
0 & \text{otherwise.}
\end{cases}
\end{equation}

The motion of moving objects is treated as a random walk,
such that they enter adjacent cells with probability $p_{out}$:
\begin{equation}
p(x_t | u_t, x_{t-1}) = \begin{cases}
1 - p_{out} &  \text{if } x_t = x_{t-1}\text{,} \\
p_{out} / 8 &  x_t \text{ is adjacent to }x_{t-1}\text{,} \\
0 & \text{otherwise.}
\end{cases}
\end{equation} 

Observations are made by detecting the objects
in images recorded by a downward-facing camera.
For each \ac{UAV}, state estimation is performed by fusing \ac{GPS}
and visual odometry data.
By combining this with attitude information,
we determine a detected object's position in a fixed coordinate frame,
and compute $p(z_t | x_t)$ given its color and position.

To reduce calculation,
we approximate Eq.~\ref{E:reward_function} as Eq.~\ref{E:reward_function_approximation}:
\begin{eqnarray}
\mathrm{R}_\mathcal{T}(a) = \sum_{i \in \mathcal{T}_a} p(i | a) \left[\mathrm{J}(\mathcal{T} + i, t - c_a) - \mathrm{J}(\mathcal{T}, t) \right] \text{,} \label{E:reward_function_approximation}
\end{eqnarray}
where
$p(i | a)$ represents the probability of finding the specific task $i \in \mathcal{T}_a$ with action $a$.

\subsubsection{Reward prediction function}
From Section~\ref{sec:RewardFunction},
reward predictions are determined
by calculating the maximum attainable reward for a set of found tasks within a time limit.
To do this, we cast Eq.~\ref{E:reward_prediction} as the well-known knapsack problem
and solve it using dynamic programming~\cite{martello1999dynamic}.

For our application, the set of found tasks contains their costs and rewards as defined in Section~\ref{ss:task_definitions}.
We treat the moving objects as static within a certain time frame,
such that their probability of moving to cells which distance is larger than the camera's viewing range is zero.
If the time since the last observation of a moving object exceeds a threshold,
it is considered unknown and must be searched for again.
However, moving objects propagate their probability, making it easier to search for them again.

A key concept is that the task cost changes with an agent's position.
If the \ac{UAV} is close to an object,
$t_{approach}$ is relatively short.
Moreover, upon a delivery,
the \ac{UAV} starts from the dropping box position.
The choice of the first object is thus addressed as a special case,
given that the order of the later objects does not change the optimization
(the \ac{UAV} always starts from the dropping box position).

We handle this by using two tables in our dynamic programming method.
The first and second tables calculate the maximum reward
with the cost considered from the dropping box position
and an arbitrary \ac{UAV} position in the arena, respectively,
corresponding to successive and first objects.
As new objects are found,
they are added to the table and assigned one of three decision labels:
(1) pick up now (first), (2) pick up later (successive), or (3) do not pick up.

The first table $B(k, \tau)$, considering only cost from the dropping box position, is calculated as:
\begin{eqnarray}
    \scalebox{0.9}{$
B(k, \tau) = \begin{cases}
\max\{B(k-1, \tau), B(k-1, \tau - c_k) + r_k\} \\ 
\qquad \qquad \qquad \qquad \qquad \text{if } c_k  \leq \tau\text{,}\\
B(k-1, \tau)\text{,}\\ 
\qquad \qquad \qquad \qquad \qquad \text{else,}
\end{cases} 
    $}
    \label{E:first_table}
\end{eqnarray}
where $c_k$ is the cost to pick up from the dropping box and $r_k$ is the reward of the task $k$.

Using Eq.~\ref{E:first_table},
the second table $B^*(k, \tau)$, considering cost from the current \ac{UAV} position, is calculated as:
\begin{equation}
    \scalebox{0.9}{$
B^*(k, \tau) = \begin{cases}
\max\{B^*(k-1, \tau), B(k-1, \tau - c^*_k) + r_k, \\ 
\qquad \qquad \qquad \qquad \ B^*(k-1, \tau - c_k) + r_k \} \\
\qquad \qquad \qquad \qquad \qquad  \text{if } c_k, c_k^*  \leq \tau\text{,}\\
\max\{B^*(k-1, \tau), B(k-1, \tau - c^*_k) + r_k\} \\
\qquad \qquad \qquad \qquad \qquad \text{if } c_k > \tau, c_k^*  \leq \tau\text{,}\\
\max\{B^*(k-1, \tau), B^*(k-1, \tau - c_k) + r_k \} \\
 \qquad \qquad \qquad \qquad \qquad \text{if } c_k  \leq \tau, c_k^* > \tau\text{,}\\
B^*(k-1, \tau) \\
 \qquad \qquad \qquad \qquad \qquad \text{else\text{,}}
\end{cases}
    $}
\end{equation}
where $c_k^*$ is the pick-up cost from the current \ac{UAV} position.
$B(k-1, \tau - c^*_k) + r_k$, $B^*(k-1, \tau - c_k) + r_k$,
and $B^*(k-1, \tau)$ correspond to decision labels (1), (2), and (3), respectively.

This reward prediction function captures
the probability of finding objects during exploration.
For example,
with little time remaining,
searching areas far away from the dropping box
offers no rewards since deliveries are impossible.

\subsection{Action Definitions}
\begin{figure*}[!ht]
 \begin{minipage}{0.20\hsize}
  \begin{center}
   \includegraphics[height=2.2cm]{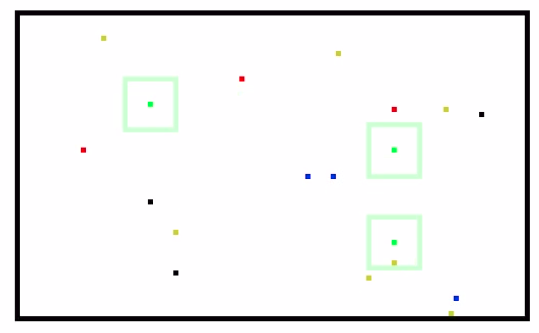}
  \end{center}
  \captionsetup{labelformat=empty,labelsep=none}
  \subcaption{2D simulator}
 \end{minipage}
 \begin{minipage}{0.28\hsize}
  \begin{center}
   \includegraphics[height=2.2cm]{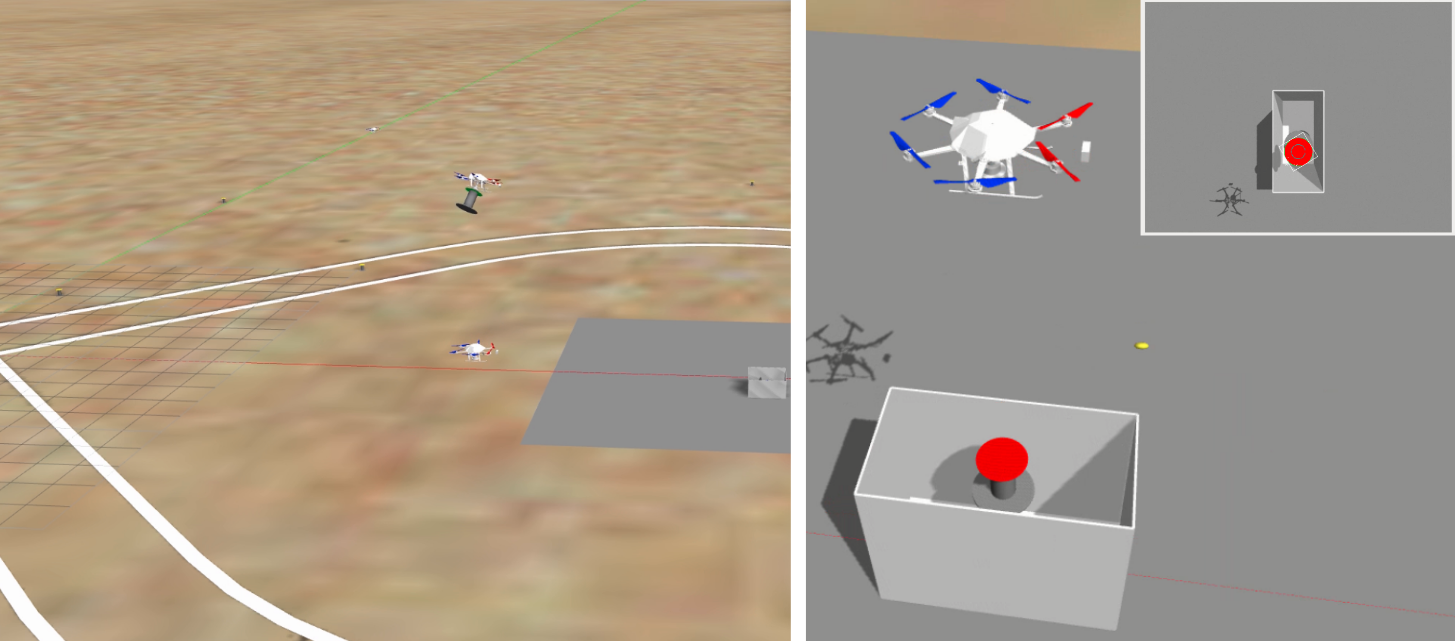}
  \end{center}
  \captionsetup{labelformat=empty,labelsep=none}
  \subcaption{3D simulator}
 \end{minipage}
 \begin{minipage}{0.24\hsize}
  \begin{center}
   \includegraphics[height=2.2cm]{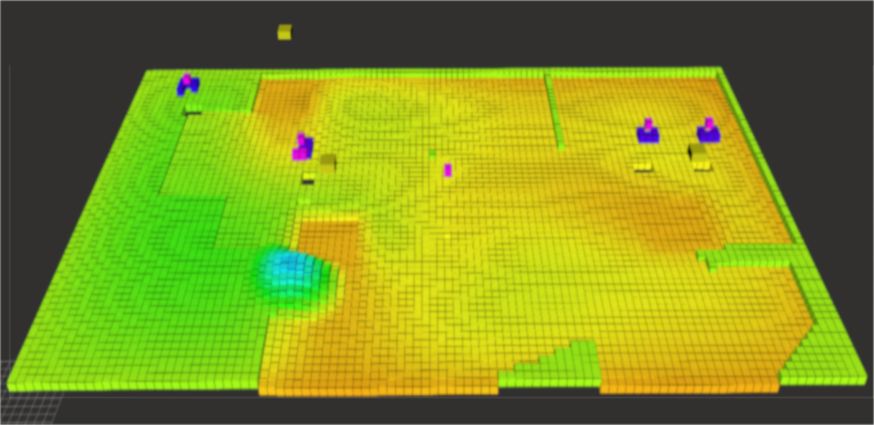}
  \end{center}
  \captionsetup{labelformat=empty,labelsep=none}
  \subcaption{\ac{PDM}}
 \end{minipage}
 \begin{minipage}{0.24\hsize}
  \begin{center}
   \includegraphics[height=2.2cm]{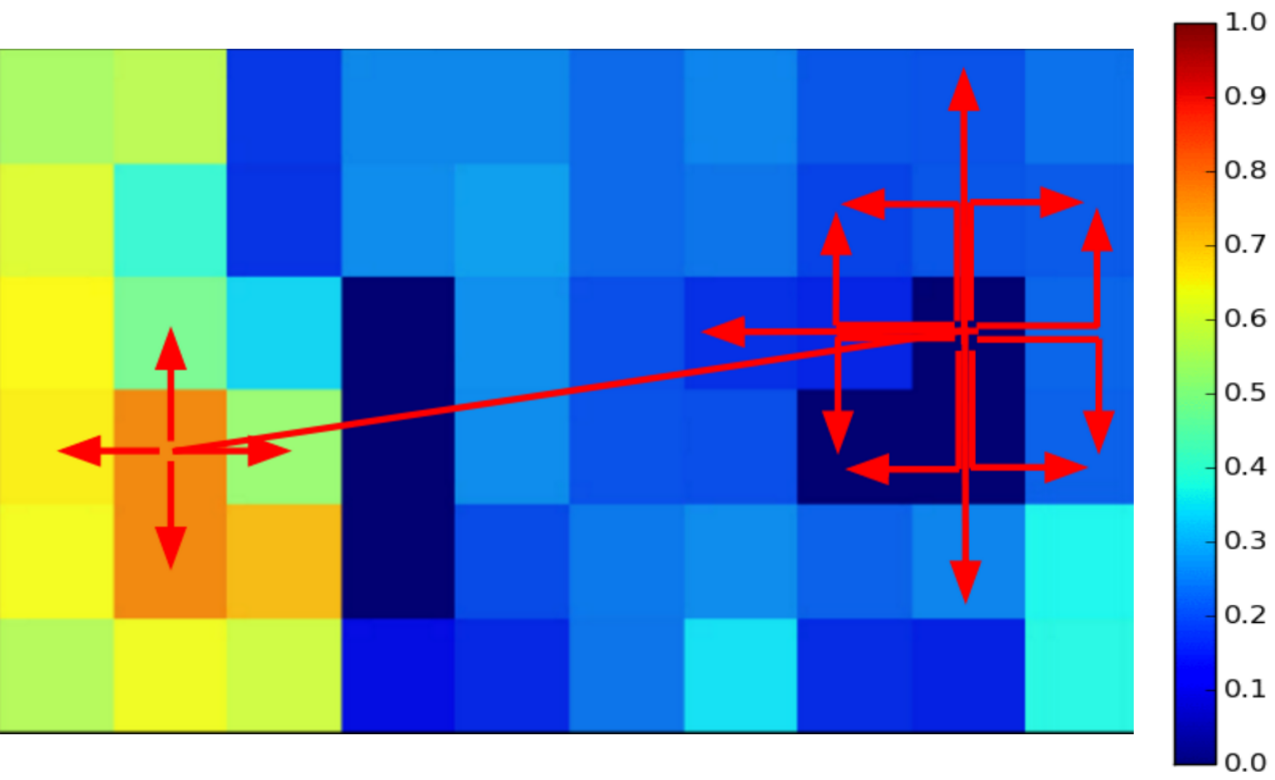}
  \end{center}
  \captionsetup{labelformat=empty,labelsep=none}
     \subcaption{\ac{PDM} grid for planning} 
  \label{fig:pdm_planner}
 \end{minipage}
 \caption{(a) shows our 2D simulator with \acp{UAV} (green), their camera \acp{FoV} (green squares), static objects (red, blue, black), and moving objects (yellow).
 (b) depicts our RotorS setup used to simulate missions with all system components.
 (c) visualizes the PDM used for predicting scores
 with the position of each \ac{UAV} (yellow cube) and its planned path (red line).
 The height of each colored cell corresponds to the probability of finding objects.
 Based on this, (d) illustrates the grid used for planning exploration paths.
 The color of a cell represents the sum of predicted score ($\sum_i p(i)r_i$, where $p(i)$ is the probability of finding object $i$ with score $r_i$).
 The color scale transitions from blue (low) to red (high).
 Red arrows indicate possible paths emanating from a \ac{UAV} position.
 We also consider paths exploring the cell with the highest predicted score, shown in orange.}
 \label{fig:experiment_setup}
\end{figure*}

A \textit{search} action (exploration) involves flying a path in the arena
to obtain measurements from the downward-facing camera.
To plan paths,
we use 2D grid-based planning on the PDM at a constant altitude,
where each cell has the dimensions of the camera \ac{FoV} (Fig.~\ref{fig:pdm_planner}). 
For an arbitrary horizon,
we enumerate every path executable from the current cell.
In addition, the paths starting from the highest-probability cell are considered
(four red arrows in Fig.~\ref{fig:pdm_planner}).
The cost of each action is computed as the travel time assuming constant velocity.

A \textit{task} action involves picking up an object, then transferring it to and depositing it in the dropping box.
This cost is defined in Eq.~\ref{func: pickup_cost}.

\section{EXPERIMENTAL SETUP}

\label{sec:experiment}
This section outlines the setups used for our simulation experiments.
First, we detail the 2D simulator used to validate our framework
by comparison to different decision-making methods.
Then, we present the 3D simulator developed
for testing the system with all mission components.

\subsection{2D Simulation} \label{ss:2d_sim}
Our decision-making framework is validated in a python-based 2D simulator.
It is assumed that the three \acp{UAV} fly at constant altitude
and detect objects within the camera \ac{FoV} reliably.
Table~\ref{tab:parameter_2d} summarizes our experimental parameters.
\begin{table}[htbp]
\centering
\caption{2D simulation parameters}
\label{tab:parameter_2d}
\begin{tabular}{lrl}
\toprule
    \textbf{Objects} & No. of 1-point static & 4     \\ 
                        &  No. of 2-point static &                                3    \\
                        & No. of 3-point static  &                                3    \\
                        &  No. of 3-point moving                                 & 10   \\
                        & Velocity of moving                & $1$\,m$/$s \\
\midrule
    \textbf{\ac{UAV}} &  Camera \ac{FoV} area                  &   $10 \times 10$\,m   \\
                        &  Velocity                       &  $2$\,m$/$s    \\ 
\midrule
    \textbf{Reward} &  Grid resolution            &   $10$\,m   \\
                        &  $t_\text{pick}$ static & $25$\,s   \\ 
                        &  $t_\text{pick}$  moving & $45$\,s   \\
                        &  $t_\text{drop}$ static   &  $20$\,s   \\ 
                        &  $t_\text{drop}$ moving & $20$\,s   \\
                        &  Calculation time          &   $10$\,s   \\
                        &  Tracking timeout for moving &   $4$\,s   \\
\bottomrule
\end{tabular}
\end{table}

\begin{figure}[htbp]
    \centering
   \includegraphics[width=0.8\linewidth]{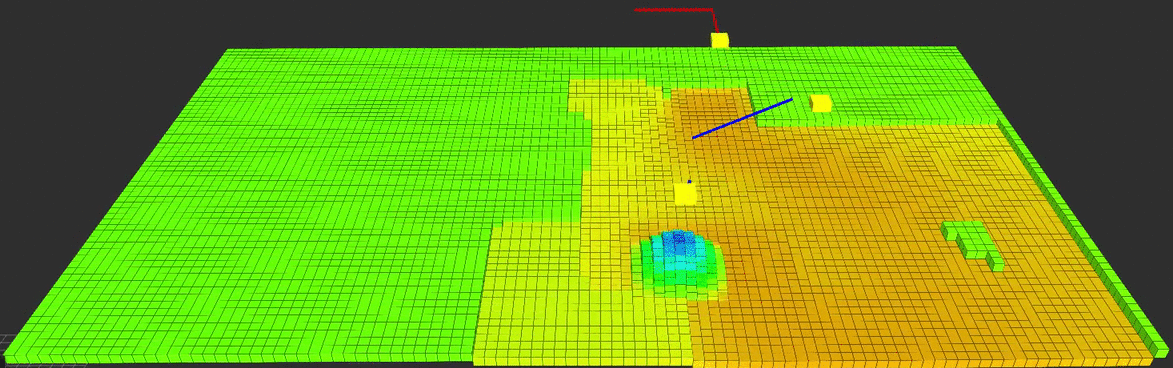}\\[2pt]
   \includegraphics[width=0.8\linewidth]{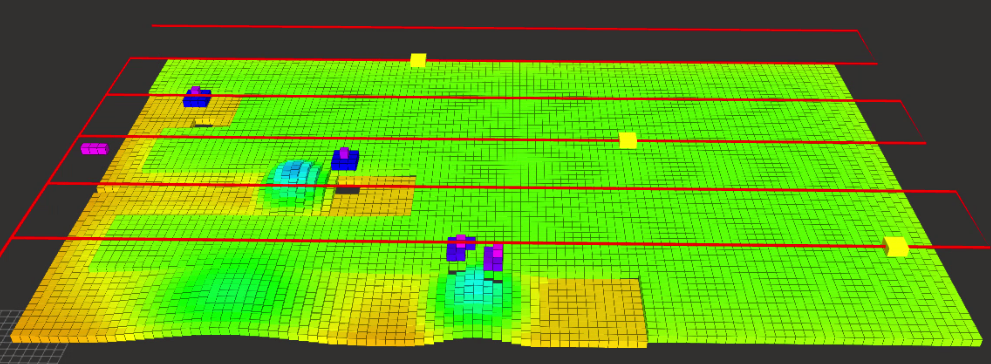}\\[2pt]
   \includegraphics[width=0.8\linewidth]{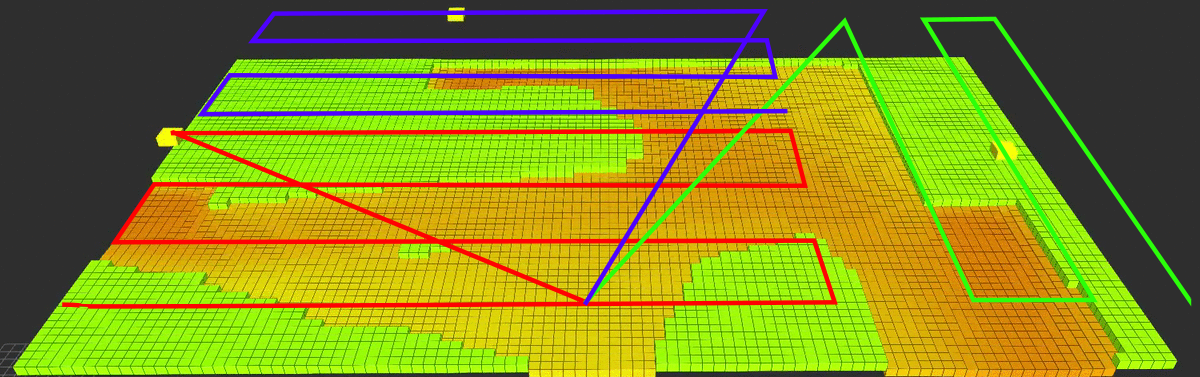}
 \caption{Three decision-making strategies used as benchmarks to evaluate our framework in 2D simulation: (top) Random, (middle) Cover-field-first, and (bottom) Cover-and-pickup.}
 \label{fig:strategies}
\end{figure}

We use this setup to evaluate our algorithm against the three different decision-making strategies illustrated in Fig.~\ref{fig:strategies} and described below.

\subsubsection{Random}
 All \acp{UAV} move by following random paths. When a \ac{UAV} finds an object, it picks it up immediately.
\subsubsection{Cover-field-first}
 All \acp{UAV} first cover the field with a ``zig-zag'' coverage pattern and then pick up the static objects found.
 The order of objects picked up is chosen based on time efficiency (the cost of an object divided by its reward).
 When a \ac{UAV} finds a moving object, it picks it up immediately.
 After picking up all the static objects, the \acp{UAV} explore the field randomly to find moving objects.
\subsubsection{Cover-and-pickup}
Each \ac{UAV} is assigned an area of the arena to cover using a ``zig-zag'' pattern.
When a \ac{UAV} finds an object, it picks it up immediately, and returns to its last position upon dropoff to continue the coverage path.
 
For each strategy above, and our own,
we consider different initial time limits $t_0 = 100, 200, \ldots,\,900$\,s
with the parameters in Table~\ref{tab:parameter_2d}.
For each budget, five trials are conducted
with randomly initialized object positions.
 
\subsection{3D Simulation} \label{ss:3d_sim}
We use RotorS~\citep{Furrer2016}, a Gazebo-based simulation environment,
to demonstrate our system running in a mission scenario.
Our \acp{UAV} are three AscTec Firefly platforms modelled
with realistic dynamics and on-board sensors.

Our decision-making unit is implemented in an integrated system
containing all other functions required for the mission,
including attitude control, localization, object detection, state machine, visual servoing for pickup, collision avoidance, etc.~\citep{bahnemann2017decentralized}
This module receives the objects detected in images
and estimated \ac{UAV} states as inputs,
and outputs actions for waypoint-based exploration or target pickup.

\subsection{System Overview}
\begin{figure}[htbp]
    \centering
   \includegraphics[width=0.9\linewidth]{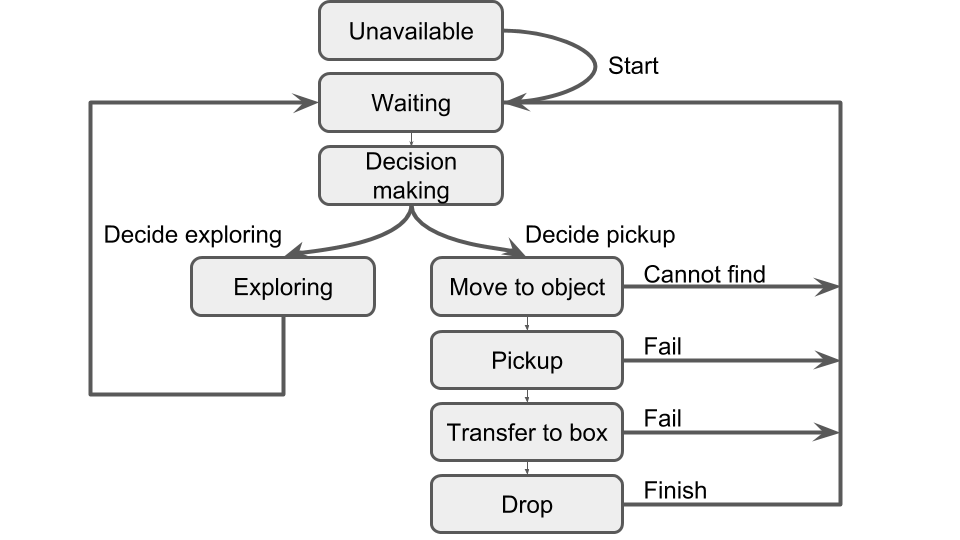}
 \caption{State machine of a \ac{UAV}. The \ac{UAV}s perform decision-making after each action (explore or pickup an object)
 while sending their decision, position, and measurements to each other.}
 \label{fig:state}
\end{figure}

Each \ac{UAV} in our system has its own PDM and shares decision and measurement information with the others.
The \acp{UAV} update the PDM individually, but using the same information, by relaying object detections to each other.
The decision proceeds as shown in Fig. \ref{fig:state}. The shared decisions are used for implicit coordination. 
In addition, the \ac{UAV}s can detect the availability of other \ac{UAV}s and can plan based on this information. This enables the agents to adapt if some agents crashed.

\section{EXPERIMENTAL RESULTS}
This section presents our experimental results.
We evaluate our framework against benchmarks in 2D simulation.
Then, we show examples of different decision-making behaviour
to highlight the benefits of our approach
in a typical mission.

\subsection{Comparison Against Benchmarks}
\begin{figure}[htbp]
    \centering
   \includegraphics[width=\linewidth]{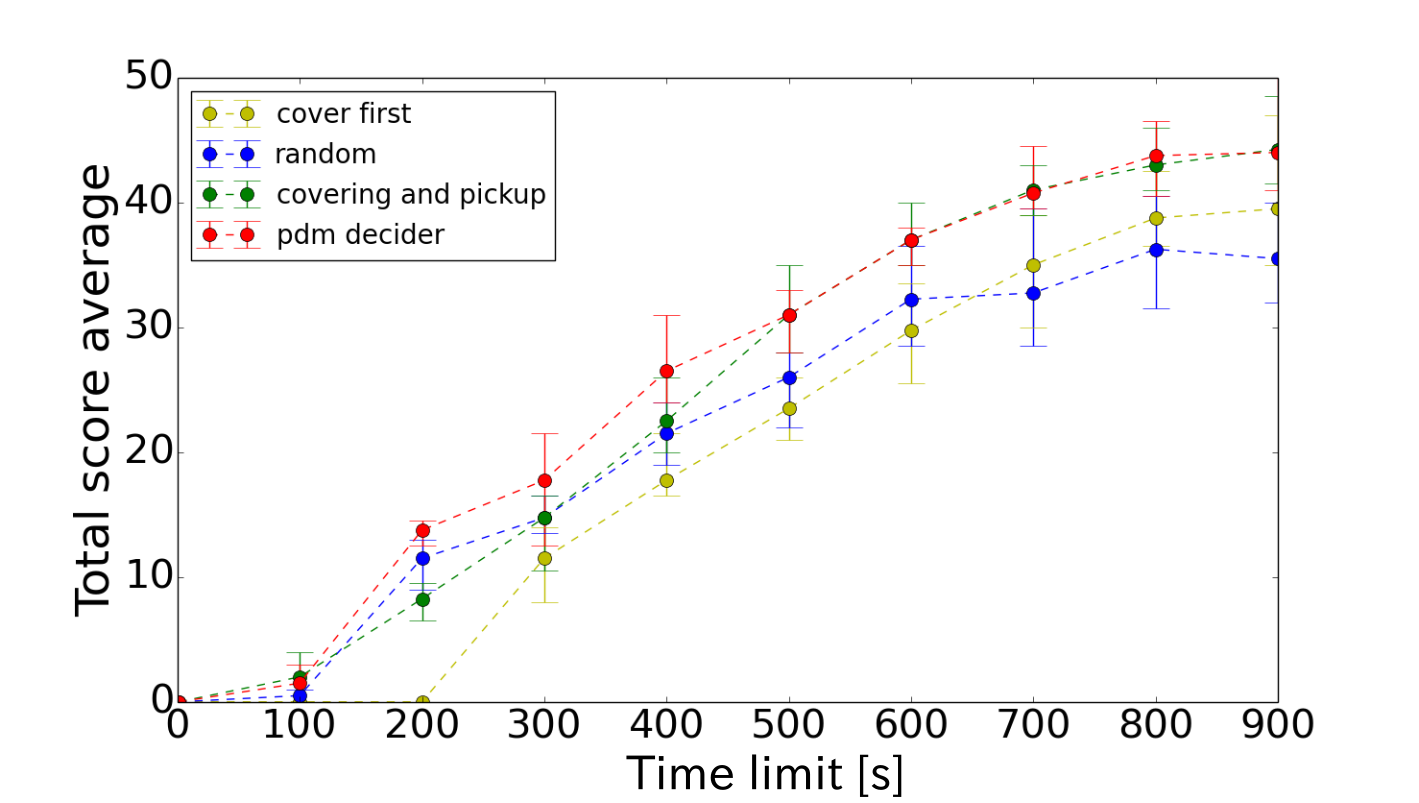}
 \caption{Comparison of our decision-making method (red) against benchmarks
 with different initial time limits.
 The solid dots show the Challenge scores averaged over 5 trials, with error bars indicating extrema.
 By accounting for the time remaining for search and action decisions,
 our method performs best across the tested range.
 }
   \label{fig:time_along}
\end{figure}

Fig.~\ref{fig:time_along} depicts the evaluation results
with varying time limits $t_0$,
as described in Section~\ref{ss:2d_sim}.
The $y$-axis measures the average competition points attained
over the five trials.
As expected,
the cover-field-first method (yellow)
obtains no points with shortest limits ($0-200$\,s)
because it lacks time to start collecting objects.
This confirms that a decision-making strategy
is needed to balance search and action
when an exhaustive search is impossible.
Our approach (red) performs best
with shorter limits ($200-400$\,s).
Unlike the benchmarks,
our algorithm uses reward predictions
to distinguish between object types given the \ac{UAV} positions,
allowing them to decide between
search and pickup within the time limit,
e.g., to seek a more valuable object
than one already found.
We show this with examples in Section~\ref{ss:decision_examples}.

With longer limits ($700-900$\,s),
the importance of decision-making decreases
as there is more time for deliveries.
The covering-and-pickup strategy (green)
scores highly since its ``zig-zag'' path ensures complete field coverage.
Our method performs competitively even without this guarantee,
as it keeps track of areas already explored.
The cover-first method is worse in comparison
as the \acp{UAV} search for highly-valued moving objects only later in the mission.
The performance of the random strategy (blue) also deteriorates
due to its low probability of finding objects away from the dropping box.

\begin{figure}[htp]
 \begin{minipage}{0.49\hsize}
  \begin{center}
   \includegraphics[width=\linewidth]{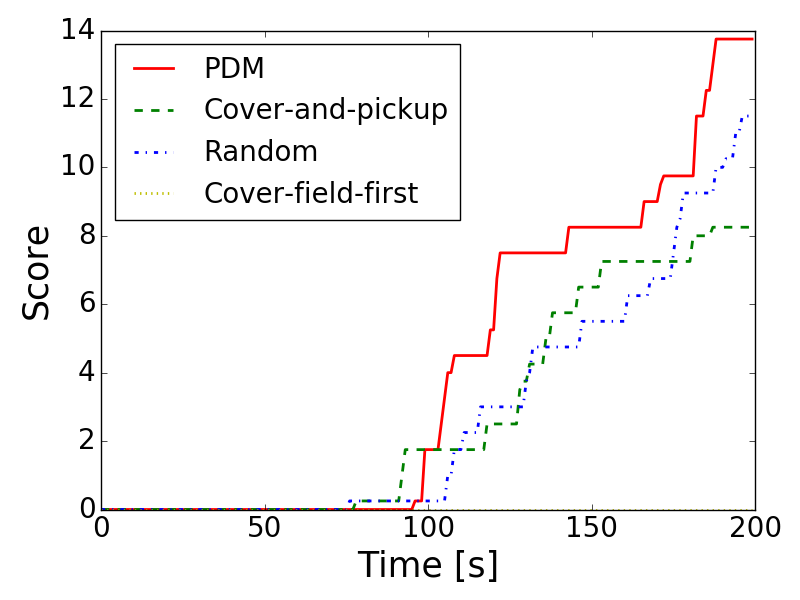}
  \end{center}
  \subcaption{Time limit: $200$\,s} \label{sf:time_along_200}
 \end{minipage}
 \begin{minipage}{0.49\hsize}
  \begin{center}
   \includegraphics[width=\linewidth]{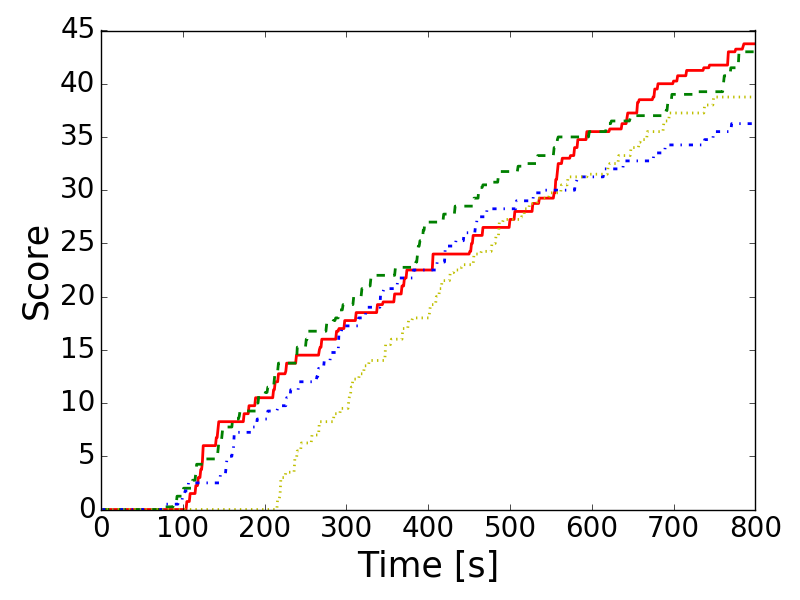}
  \end{center}
  \subcaption{Time limit: $800$\,s} \label{sf:time_along_800}
  \label{fig:pdm_initial}
 \end{minipage}
 \caption{Comparison of the score point evolution for our decision-making method (red)
 against benchmarks in (a) $200$\,s and (b) $800$\,s missions.
 In (a),
 significantly more points are obtained
 using our approach
 as it accounts for the time restriction.}\label{fig:comparison}
 \end{figure}

Fig.~\ref{fig:comparison} shows the evolution of score for two missions with different limits.
Using our algorithm,
more points are obtained at later stages
as the \acp{UAV} decide to secure object pickups
when there is little time remaining.

\begin{figure*}[t]
\centering
 \begin{minipage}{0.6\hsize}
  \begin{center}
   \includegraphics[height=2.20cm]{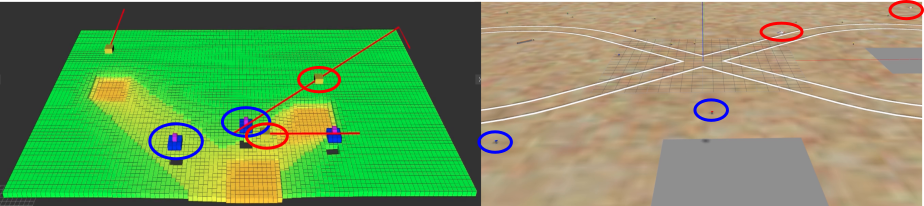}
  \end{center}
  \subcaption{Explore}
  \label{fig:choose_explore}
 \end{minipage}
 \begin{minipage}{0.39\hsize}
  \begin{center}
   \includegraphics[height=2.20cm]{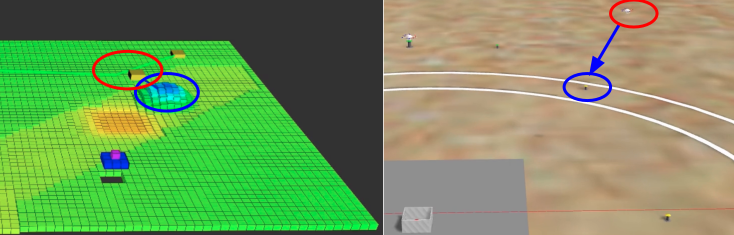}
  \end{center}
  \subcaption{Pick up moving object}
  \label{fig:pickup_moving}
 \end{minipage}
 
 \begin{minipage}{0.4\hsize}
 \vspace{2mm}
  \begin{center}
   \includegraphics[height=2.35cm]{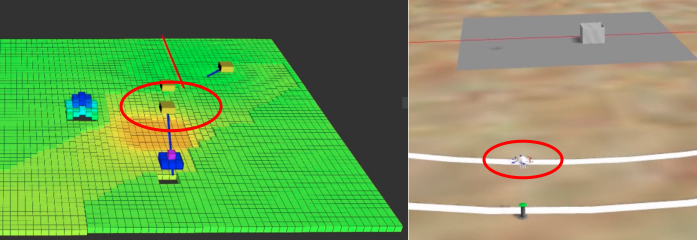}
  \end{center}
  \subcaption{Pick up static objects}
  \label{fig:pickup_stable}
 \end{minipage}
 \begin{minipage}{0.4\hsize}
 \vspace{2mm}
  \begin{center}
   \includegraphics[height=2.35cm]{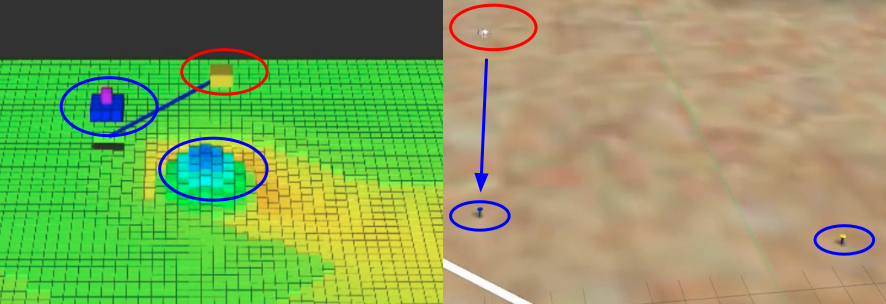}
  \end{center}
  \subcaption{Pick up static object instead of a moving one}
  \label{fig:choose_stable}
 \end{minipage}
 \caption{Examples of decision-making situations
 in our Gazebo-based environment.
 (a)-(d) show the visualization (left) and simulation views (right)
 corresponding to the four decisions described in the text.
 Blue (1pt), green (2pt), and red (3pt) shapes indicate static objects,
 while yellow (3pt) shapes represent moving ones.
 The \ac{UAV} (red circles) and object (blue circles) positions are shown.
 Red and blue signify decisions to fly exploratory paths and pick up objects, respectively,
 with blue arrows illustrating pick-up actions.
 By using a reward prediction function,
 our algorithm accounts for different trade-off aspects
 to maximize the score obtained.}
\end{figure*}
\subsection{Decision-making Examples in a 3D Simulation} \label{ss:decision_examples}

In the following, we present examples of decision-making
using our framework
in the RotorS environment (Section~\ref{ss:3d_sim}).
The aim is to show that our method
can account for different situational aspects when making trade-offs,
e.g., picking up moving objects before they are lost,
choosing between search and action based on the time remaining.
 
\begin{enumerate}[label=(\alph*)]
 \item \textit{Explore}.
Early in the mission,
the \acp{UAV} decided not to pick up objects despite having found them,
as there was ample time to search for alternatives (Fig~\ref{fig:choose_explore}).
Exploratory paths with implicit coordination are generated
to maximize the predicted total score based on the \ac{PDM}.

\item \textit{Pick up moving object}.
A \ac{UAV} picked up a moving object upon its detection (Fig.~\ref{fig:pickup_moving}).
Because moving objects are considered static only temporarily,
neglecting pickup
would lead to the object becoming unknown in the future,
causing the predicted total score to decrease.

\item \textit{Pick up static objects}.
With many static objects and little time remaining,
the \acp{UAV} decided to pick the objects up
(Fig.~\ref{fig:pickup_stable}).
As there was insufficient time to both risk exploration
and return to the objects later,
a decision to search caused the predicted total score to decrease.

\item \textit{Pick up static object instead of a moving one}.
A \ac{UAV} found both a new static and a new moving object (Fig.~\ref{fig:choose_stable}).
With enough time,
the moving object would be targeted.
However, in a situation with little time remaining,
the static object is preferred because its pick-up time is shorter.
This decision considers both the time constraints and object type,
so
cannot be performed with simple rule-based logic, e.g., ``Pick up a moving object as soon as it is found".
\end{enumerate}

\section{CONCLUSION}
\label{sec:conclusion}
This work introduced a decentralized multi-agent decision-making framework
for the search and action problem in a time-constrained setting.
Our algorithm uses probabilistic reasoning to make decisions
leading to highest predicted rewards over a given time limit.
The near-optimality of the output policies is guaranteed
by exploiting the properties of the optimization function
used for implicit coordination.

Our framework was applied
in the MBZIRC search, pick, and place scenario
by specifying a \ac{PDM} and reward prediction function for action selection.
We showed its advantages over alternative decision-making strategies
in terms of mission performance with varying time limits.
Experiments in a 3D environment
demonstrated real-time system integration 
with examples of informed decision-making.

Future research will address adapting our algorithm
to different scenarios, e.g. search-and-rescue.
We aim to apply our algorithm for further search-and-action scenarios where the time cost and reward can be defined for all tasks and exploring actions. Here, the time limit can reflect a limited resource, e.g., battery level or energy consumptions.
 Another interesting avenue for future research is the use of other methods than bayesian filtering and PDM for the probability calculation of finding new tasks, e.g., for unknown field size and probability distribution.
Possible extensions involve allowing for flight at variable altitudes,
alternative sensor types, and unknown environments/tasks.

\section*{ACKNOWLEDGMENT}
This project has received funding from the European Union’s Horizon 2020 
research and innovation programme under grant agreement No 644227 and from the 
Swiss State Secretariat for Education, Research and Innovation (SERI) under 
contract number 15.0029,
and was partially sponsored by the Mohamed Bin Zayed International Robotics Challenge.


\appendix
\addtolength{\textheight}{-0cm}
In the following, we sketch the proof of the aforementioned properties (positive monotonicity and submodularity) of the optimization function specified in Eq~\ref{E:optimization_function}.
\subsection{Positive monotonicity}
If the function $\mathrm{F}$ satisfies:
\begin{eqnarray}
A \subseteq B\Rightarrow
\mathrm{F}(A) \leq \mathrm{F}(B)\mathrm{,}
\end{eqnarray}
it is called non-decreasing.

\textit{Proof}. $\mathrm{F}(A) = \mathrm{J}(\mathcal{T}_A, \tau_0)$, and $\mathcal{T}_A \subseteq \mathcal{T}_B$ because $B$ contains all actions in $A$. From the definition of $\mathrm{J}(\mathcal{T}, \tau)$,
it is obvious that with a larger set of tasks, $\mathrm{J}(\mathcal{T}, \tau)$ will get larger.
As $\mathrm{F}$ satisfies this condition, it is proven to be nondecreasing.

\subsection{Submodularity}
If the function $\mathrm{F}$ satisfies:
\begin{eqnarray}
A \subseteq B \Rightarrow
\mathrm{F}(A\cup e) - \mathrm{F}(A) \geq \mathrm{F}(B \cup e) - \mathrm{F}(B)\mathrm{,}
\end{eqnarray}
it is called submodular.

This means that when a new set $e$ is added, the increase of $\mathrm{F}$ is smaller with the larger set $B$ than the smaller set $A$.

\textit{Proof}. Since $\mathrm{F}(A) = \mathrm{J}(\mathcal{T}_A, \tau_0)$,
there are three cases, distinguished by whether the total cost of tasks exceeds the time limit.

\textbf{Case 1}. $\sum_{i \in A}c_i + c_e \leq \tau_0$ and $\sum_{i \in B}c_i + c_e \leq \tau_0\textrm{.}$

In this case, the reward increase is as same as the reward $r_e$ of task $e$.

As a result,
\begin{equation}
\mathrm{F}(A\cup e) - \mathrm{F}(A) = r_e =  \mathrm{F}(B \cup e) - \mathrm{F}(B)\textrm{.}
\end{equation}

\textbf{Case 2}. $\sum_{i \in A}c_i + c_e \leq \tau_0$ and $\sum_{i \in B}c_i + c_e > \tau_0\textrm{.}$

In this case, $\mathrm{F}(B\cup e) - \mathrm{F}(B) \leq r_e$ because the tasks must be selected to satisfy the constraint, which would reduce the increase of $\mathrm{F}$.

Therefore,
\begin{equation}
\mathrm{F}(A\cup e) - \mathrm{F}(A) = r_e \geq  \mathrm{F}(B \cup e) - \mathrm{F}(B)
\end{equation}

\textbf{Case 3}. $\sum_{i \in A}c_i + c_e > \tau_0$ and $\sum_{i \in B}c_i + c_e > \tau_0\textrm{.}$
This means that the tasks are already selected from $A$ to match the constraint.
Now, consider adding a new task $e$.
Let $\sum_{i \in A} c_ix_i = \tau_A \leq \tau_0$.\\
In this case, a reward increase occurs by replacing tasks $j$ with $e$:

\begin{eqnarray}
\max_{j \in \mathcal{T}_A}(r_e - \sum r_j)\textrm{,} \\
\text{with}\  \tau_A + c_e - \sum c_j \leq \tau_0\textrm{.}
\end{eqnarray}

With a larger set of tasks $B$,
the tasks initially selected are gained by replacing the tasks from $A$.
If tasks $j$ are not replaced in this operation,
the reward increase is the same.
If the tasks are already replaced, the reward increase is smaller than $ r_e - \sum r_j $ because they are already replaced by more valuable tasks. Then,
\begin{equation}
\mathrm{F}(A\cup e) - \mathrm{F}(A) = r_e - \sum r_j \geq  \mathrm{F}(B \cup e) - \mathrm{F}(B)\textrm{.}
\end{equation}

As $\mathrm{F}$ satisfies the three cases above, it is proven to be submodular.
 \label{sec:proof}
\bibliographystyle{IEEEtranN}
\footnotesize
\bibliography{references/bibliography}

\end{document}